\documentclass[10pt,logo,copyright]{giga-report}
\usepackage[authoryear,sort&compress,round]{natbib}

\usepackage[utf8]{inputenc} %
\usepackage[T1]{fontenc}    %

\usepackage{parskip}        %
\usepackage{url}            %
\usepackage{booktabs}       %
\usepackage{amsfonts}       %
\usepackage{nicefrac}       %
\usepackage{microtype}      %
\usepackage{xcolor}         %
\usepackage[dvipsnames]{xcolor} %
\usepackage{graphicx}
\usepackage{animate}        %
\usepackage{subcaption}
\usepackage{tabularx}
\usepackage{makecell}
\usepackage{adjustbox}
\usepackage{setspace}
\usepackage{todonotes}
\usepackage{colortbl}
\usepackage{wrapfig}

\newcolumntype{M}[1]{>{\centering\arraybackslash}m{#1}}
\usepackage{float}
\usepackage{tikz}
\usetikzlibrary{positioning,shapes,arrows}
\usepackage{amsmath,amsfonts,bm, bbm,leftindex}
\usepackage{multirow}
\usepackage{comment}
\usepackage{gensymb}
\usepackage{lipsum}
\usetikzlibrary{arrows.meta, positioning, fit}
\usepackage[para]{threeparttable}
\usepackage{tikz}
\usetikzlibrary{tikzmark}

\def\eqref#1{equation~\ref{#1}}

\def\1{\bm{1}}

\DeclareMathAlphabet{\mathsfit}{\encodingdefault}{\sfdefault}{m}{sl}
\SetMathAlphabet{\mathsfit}{bold}{\encodingdefault}{\sfdefault}{bx}{n}

\makeatletter
\let\save@mathaccent\mathaccent
\newcommand*\if@single[3]{%
  \setbox0\hbox{${\mathaccent"0362{#1}}^H$}%
  \setbox2\hbox{${\mathaccent"0362{\kern0pt#1}}^H$}%
  \ifdim\ht0=\ht2 #3\else #2\fi
  }
\newcommand*\rel@kern[1]{\kern#1\dimexpr\macc@kerna}
\newcommand*\widebar[1]{\@ifnextchar^{{\wide@bar{#1}{0}}}{\wide@bar{#1}{1}}}
\newcommand*\wide@bar[2]{\if@single{#1}{\wide@bar@{#1}{#2}{1}}{\wide@bar@{#1}{#2}{2}}}
\newcommand*\wide@bar@[3]{%
  \begingroup
  \def\mathaccent##1##2{%
    \let\mathaccent\save@mathaccent
    \if#32 \let\macc@nucleus\first@char \fi
    \setbox\z@\hbox{$\macc@style{\macc@nucleus}_{}$}%
    \setbox\tw@\hbox{$\macc@style{\macc@nucleus}{}_{}$}%
    \dimen@\wd\tw@
    \advance\dimen@-\wd\z@
    \divide\dimen@ 3
    \@tempdima\wd\tw@
    \advance\@tempdima-\scriptspace
    \divide\@tempdima 10
    \advance\dimen@-\@tempdima
    \ifdim\dimen@>\z@ \dimen@0pt\fi
    \rel@kern{0.6}\kern-\dimen@
    \if#31
      \overline{\rel@kern{-0.6}\kern\dimen@\macc@nucleus\rel@kern{0.4}\kern\dimen@}%
      \advance\dimen@0.4\dimexpr\macc@kerna
      \let\final@kern#2%
      \ifdim\dimen@<\z@ \let\final@kern1\fi
      \if\final@kern1 \kern-\dimen@\fi
    \else
      \overline{\rel@kern{-0.6}\kern\dimen@#1}%
    \fi
  }%
  \macc@depth\@ne
  \let\math@bgroup\@empty \let\math@egroup\macc@set@skewchar
  \mathsurround\z@ \frozen@everymath{\mathgroup\macc@group\relax}%
  \macc@set@skewchar\relax
  \let\mathaccentV\macc@nested@a
  \if#31
    \macc@nested@a\relax111{#1}%
  \else
    \def\gobble@till@marker##1\endmarker{}%
    \futurelet\first@char\gobble@till@marker#1\endmarker
    \ifcat\noexpand\first@char A\else
      \def\first@char{}%
    \fi
    \macc@nested@a\relax111{\first@char}%
  \fi
  \endgroup
}
\makeatother

\definecolor{darkred}{rgb}{0.7, 0.0, 0.0}

\usepackage{amsmath} 

\usepackage[table]{xcolor}
\usepackage{pifont}
\usepackage{graphicx}

\usepackage[nameinlink]{cleveref}
\crefname{equation}{Eq.}{Eqs.}
\crefname{figure}{Fig.}{Figs.}
\crefname{section}{Sec.}{Sec.}
\crefname{appendix}{App.}{App.}
\crefname{table}{Tab.}{Tabs.}
\crefname{algorithm}{Algo}{Algo}
\crefname{thm}{Thm}{Thm}
\Crefname{thm}{Thm}{Thm}
\crefname{prop}{Prop}{Prop}

\newcommand{\crefnames}[3]{%
  \@for\next:=#1\do{%
    \expandafter\crefname\expandafter{\next}{#2}{#3}%
  }%
}

\renewcommand{\paragraph}[1]{\vspace{1.25mm}\noindent\textbf{#1}}
\newcommand{\tablestyle}[2]{\setlength{\tabcolsep}{#1}\renewcommand{\arraystretch}{#2}\centering\footnotesize}
\renewcommand{\paragraph}[1]{\vspace{1.25mm}\noindent\textbf{#1}}
\newlength\savewidth

\definecolor{degray}{gray}{.6}

\title{ViVa: A Video-Generative Value Model for Robot Reinforcement Learning}

\author{%
\centering
\textbf{Jindi Lv$^{1,2\ast}$}, 
\textbf{Hao Li$^{1\ast}$},
\textbf{Jie Li$^{1}$}, 
Fankun Kong$^{1}$,
Yang Wang$^{1}$,
Pengfei Yi$^{1}$,
Yifei Nie$^{1}$, 
Xiaofeng Wang$^{1,3}$,
Zheng Zhu$^{1\dagger}$,
Chaojun Ni$^{1}$,
Qiuping Deng$^{1}$,
Hengtao Li$^{1}$,
Jiancheng Lv$^{2\dagger}$,
Guan Huang$^{1}$ \\
$^{1}$GigaAI\;
$^{2}$Sichuan University\;
$^{3}$Tsinghua University\\
$^{\ast}$ Equal Contribution \quad
$^{\dagger}$ Corresponding Authors \\
{Project Page: \href{https://viva-value-model.github.io/}{https://viva-value-model.github.io/}} \\
} 

\begin{document}
\maketitle

\begin{center}
    \vspace{15pt}
    \centering
    \captionsetup{type=figure, justification=justified, singlelinecheck=false}
    \includegraphics[width=1.0\linewidth]{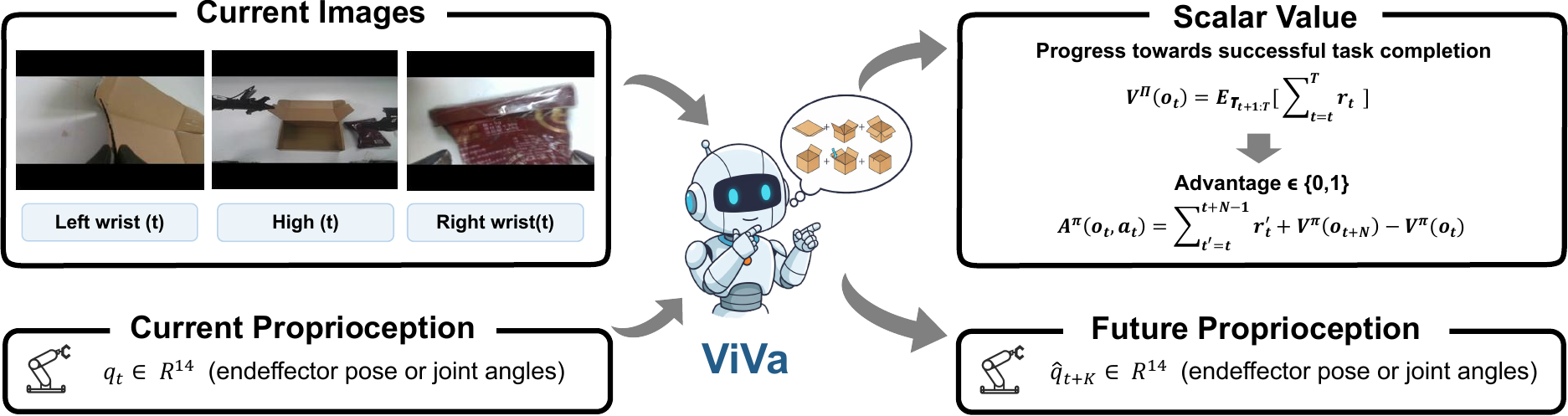}
    \caption{Illustration of ViVa. Given multi-view observations and proprioception, ViVa jointly predicts future proprioception and a task-progress value. By grounding value estimation in anticipated embodiment dynamics, it leverages spatiotemporal priors to couple value with foresight.}
    \label{fig:illustrative_framework}
\end{center}


\begin{abstract}
Vision-language-action (VLA) models have advanced robot manipulation through large-scale pretraining, but real-world deployment remains challenging due to partial observability and delayed feedback. Reinforcement learning addresses this via value functions, which assess task progress and guide policy improvement. However, existing value models built on vision-language models (VLMs) struggle to capture temporal dynamics and physical interactions, undermining reliable value estimation in long-horizon tasks.
In this paper, we propose ViVa, a video-generative value model that repurposes a pretrained video generator to jointly predict future proprioception and a scalar value. By grounding value estimation in anticipated embodiment dynamics, ViVa leverages spatiotemporal priors to intrinsically couple value with foresight beyond static snapshots. ViVa achieves state-of-the-art results in metric-based evaluation across three tasks, producing reliable value signals that accurately track task progress and detect execution errors. Integrated into RECAP, it achieves an average success rate of 80\%, highlighting the promise of video-generative models for value estimation.
\end{abstract}


\abscontent
\section{Introduction}

   Building robots that can perceive, reason, and act in the physical world remains a central challenge in embodied artificial intelligence~\citep{sapkota2025vision,li2026matters}. Vision-language-action (VLA) models~\citep{zitkovich2023rt,kim2024openvla,intelligence2025pi,li2025cogvla,team2026gigabrain} have made significant strides by leveraging large-scale pretraining to enable general-purpose manipulation across diverse tasks. Yet success in real-world settings requires more than static scene understanding: robotic interaction unfolds under partial observability and delayed feedback, where the consequences of decisions only manifest over extended horizons~\citep{huang2022language,zitkovich2023rt}. Learning to connect present behavior with future outcomes thus remains a fundamental challenge for real-world robotics.

This challenge calls for an ability that assesses whether ongoing interaction is progressing toward successful task completion. Such progress awareness allows robots to distinguish beneficial behaviors from undesirable ones and improve through experience. In reinforcement learning (RL)~\citep{sutton1998reinforcement}, this capability is formalized by the value function, which estimates expected future outcomes and provides a learning signal for policy improvement. Recent VLA frameworks such as $\pi^{*}_{0.6}$~\citep{intelligence2025pi0.6} highlight this importance: their RL with Experience and Corrections via Advantage-conditioned Policies (RECAP) pipeline relies on a value function for advantage estimation and policy refinement, demonstrating that learning performance strongly depends on value model quality.

Motivated by this importance, recent works have explored leveraging vision-language models (VLMs)~\citep{chen2024spatialvlm, comanici2025gemini,bai2025qwen3,zhu2025internvl3,li2024llava,marafioti2025smolvlm} for value estimation, framing value prediction as classification~\citep{intelligence2025pi0.6} or temporal ordering problems~\citep{ma2024vision}. While promising, these approaches inherit a key limitation: VLMs are primarily trained on static image–text data for semantic understanding rather than explicitly modeling how scenes evolve over time. Accordingly, they capture what is present in a scene but struggle to represent how interactions dynamically transform the environment. This mismatch limits their ability to support reliable value estimation in temporally extended robotic tasks.

The above limitations reveal a key insight: \textit{value estimation is inherently a problem of anticipating how the future will unfold}. In contrast to discriminative models trained on static data, video generative models are explicitly optimized to capture temporal evolution, learning how scenes change as interactions unfold. This makes them a natural foundation for value estimation, as the ability to imagine future outcomes directly enables assessing whether current behavior progresses toward task completion.  Guided by this observation, we reformulate value learning as future prediction and develop a video-generative value model.

In this paper, we propose \textbf{Vi}deo-generative \textbf{Va}lue model (\textbf{ViVa}), a novel approach that repurposes a pretrained video generator as a value function for robotic reinforcement learning. By leveraging the spatiotemporal priors learned from large-scale video corpora, our model captures rich dynamics about how scenes evolve over time. Taking the current observation together with robot proprioception as input, ViVa jointly predicts future proprioception and a scalar value for the current state (Figure~\ref{fig:illustrative_framework}). Grounding value estimation in anticipated embodiment dynamics enables ViVa to incorporate predictive structure beyond static snapshots, intrinsically coupling value with foresight. This design provides more reliable value signals for advantage computation, leading to improved policy optimization in robotic manipulation tasks.

We conduct extensive experiments across three long-horizon manipulation tasks. ViVa demonstrates strong sensitivity to fine-grained execution errors and robust generalization to novel objects. Integrated into RECAP for real-robot evaluation, it delivers consistent performance gains, highlighting the potential of video-generative models as a new paradigm for value estimation in robotics.

We highlight the main contributions of this paper below:
\begin{itemize}
    \item We identify robotic value estimation as fundamentally a future anticipation problem, for which video generative models offer a more natural foundation than discriminative VLMs.
    \item We introduce ViVa, a video-generative value model that intrinsically couples value with foresight by jointly predicting future embodiment dynamics alongside the current value.
    \item ViVa achieves state-of-the-art performance across three long-horizon tasks in metric-based evaluation and real-robot experiments, with sensitivity to execution errors and robust generalization to novel objects.
\end{itemize}
\section{Related Works}

\subsection{Value Functions in Robot Learning}

Value functions are central to reinforcement learning for robotics, providing learning signals under delayed and sparse feedback~\citep{sutton1998reinforcement,ross2011reduction}. Prior works have applied reinforcement learning to robotic manipulation, spanning offline Q-learning~\citep{kalashnikov2018scalable,levine2020offline,mandlekar2020iris,huang2025co}, autonomous real-world interaction~\citep{sharma2023self,mendonca2023alan,luo2024serl,lampe2024mastering}, and end-to-end value-guided policy optimization~\citep{zhai2025vision,ghasemipour2025self}. To enable reinforcement learning for VLA models~\citep{team2024octo,o2024open,kim2024openvla,li2024cogact,cheang2024gr,liu2024rdt}, recent works have explored VLM-based value estimation~\citep{frans2025diffusion,ma2024vision,ma2023liv}. For instance, GVL~\citep{ma2024vision} directly queries VLMs to assess task progress by framing value prediction as temporal ordering over shuffled video frames. TopReward~\citep{chen2026topreward} instead leverages VLM token probabilities as zero-shot reward signals. $\pi^{*}_{0.6}$~\citep{intelligence2025pi0.6} trains a dedicated VLM-based value model with dense supervision as a core component of the RECAP pipeline for advantage-conditioned policy refinement.


Despite their differing designs, these methods share a fundamental limitation: they all rely on VLMs trained on static images, which capture per-frame semantics but lack explicit modeling of temporal dynamics and physical interactions. This motivates leveraging video generative models, which learn spatiotemporal dynamics directly from large-scale video data and offer a natural foundation for value estimation in long-horizon tasks.

\subsection{Video Generation Models for Robot Manipulation}

Video generation models~\citep{zheng2024open,yang2024cogvideox,kong2024hunyuanvideo,blattmann2023stable} are explicitly optimized to capture temporal evolution, commonly via diffusion Transformers~\citep{peebles2023scalable,bao2023all} conditioned on language~\citep{singer2022make,villegas2022phenaki,blattmann2023stable} or visual~\citep{ceylan2023pix2video,qi2023diffdance} inputs. These properties make them well-suited for anticipating visual dynamics.

In robotics, video generation has been applied to planning via world models~\citep{zhou2024robodreamer}, where generated futures simulate action outcomes for decision making. It has also been integrated into policy learning, by extracting actions via inverse dynamics~\citep{du2023learning,yang2023learning}, conditioning policies on goal frames~\citep{du2023video,zhang2025gevrm}, or jointly generating video and actions~\citep{cheang2024gr,wu2023unleashing,ye2026world,ye2026gigaworld}. More recently, human-to-robot transfer has been explored by synthesizing human-object interaction videos~\citep{bharadhwaj2024gen2act,zhao2025taste,kareer2025emergence}.

Despite these advances, existing methods primarily leverage video generation to produce or guide actions. In contrast, we investigate a complementary role: value estimation. We propose ViVa, a video-generative value model that repurposes a pretrained video generator to predict scalar values, grounding value estimation in anticipated embodiment dynamics.

\section{Method}

\begin{figure}
    \centering

    \captionsetup{type=figure, justification=justified, singlelinecheck=false}
    \includegraphics[width=1.0\linewidth]{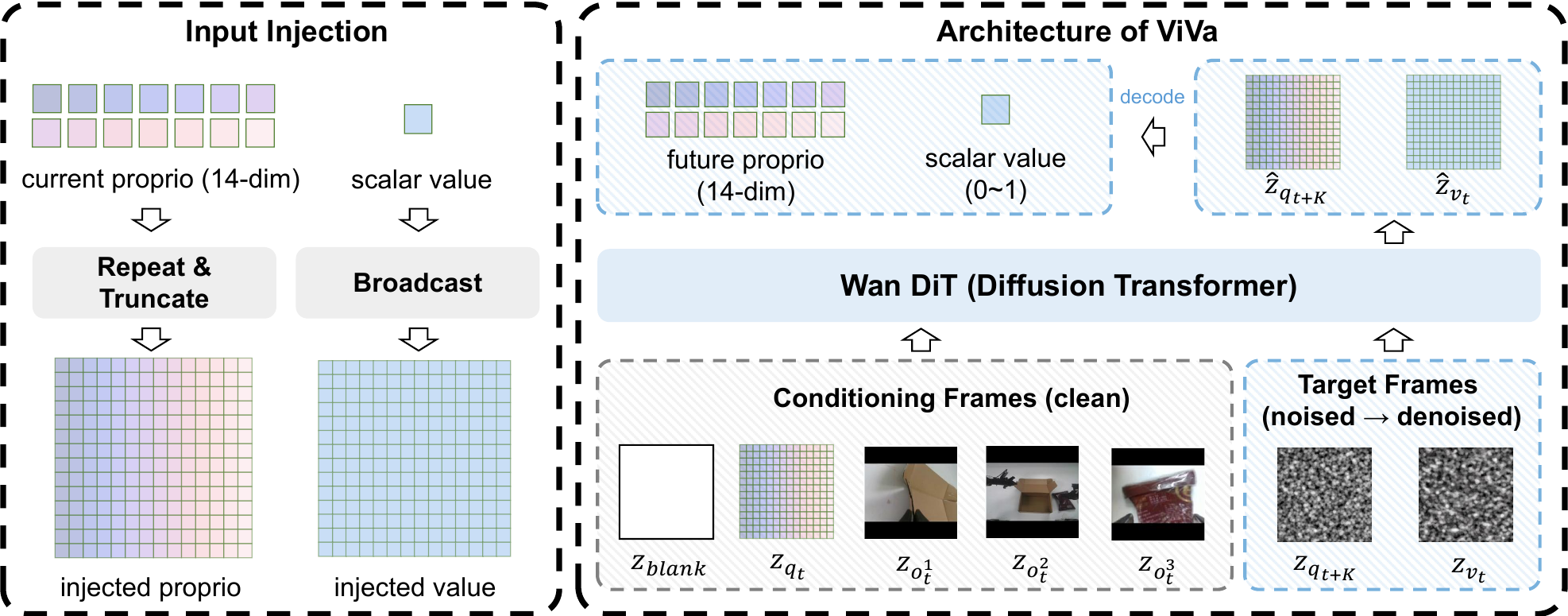}
    \caption{Overall architecture of ViVa. Left: Proprioception and the scalar value are embedded into latent frames via repeat-padding and broadcast. Right: The latent sequence comprises clean conditioning frames (blank token, proprioception, multi-view images) and noisy target frames (future proprioception and value). The diffusion Transformer denoises the targets conditioned on the clean prefix, jointly predicting future embodiment state and value.}
    \label{fig:viva_architecture}
    \vspace{-10pt}
\end{figure}
\subsection{Problem Formulation}

We formalize robotic manipulation as a Markov decision process. At each time step $t$, the agent receives an observation comprising multi-view RGB images $\mathbf{o}_t$ and proprioception $\mathbf{q}_t$; we denote the joint observation as $\mathbf{x}_t = (\mathbf{o}_t, \mathbf{q}_t)$. The value function
\[
V^{\pi}(\mathbf{x}_t) = \mathbb{E}_{\pi}\Bigl[ \sum_{k=t}^{T} r_k \;\big|\; \mathbf{x}_t \Bigr]
\]
estimates the expected future return under policy $\pi$. We argue that accurate value estimation should be grounded in understanding temporal dynamics and physical interactions, which static image-based models struggle to capture. To address this, we propose a video-generative value model that leverages spatiotemporal priors from video data, jointly predicting future proprioception alongside the current value to ground value estimation in anticipated physical dynamics.

\subsection{Overall Architecture}
\label{sec:overall_arch}
We build our video-generative value model upon Wan2.2~\citep{wan2025wan}, a pretrained video diffusion Transformer that originally generates future frames conditioned on an initial image and text. To adapt it for value estimation, we extend its input and output modalities via latent injection~\citep{agarwal2025cosmos,liang2025video}, without modifying the core architecture. The overall architecture of ViVa is illustrated in Figure~\ref{fig:viva_architecture}.

\paragraph{Latent encoding of modalities.} All input and output modalities are mapped to latent frames of shape \((H',W',C')\), where \(H',W'\) are the spatial dimensions after VAE downsampling and \(C'\) is the latent channel dimension. We use a pretrained spatiotemporal VAE to encode images: each camera view \(\mathbf{o}_t^i\) is independently compressed into a latent frame \(\mathbf{z}_{\mathbf{o}_t^i}\). For low-dimensional vectors such as the proprioceptive state \(\mathbf{q}_t\) and the scalar value \(v_t\), we design specialized injection procedures. Both are first normalized to \([-1,1]\) to match the latent space statistics. The proprioceptive state \(\mathbf{q}_t\) is embedded via repeat-padding: we repeat its elements to match the latent frame size \(H'W'C'\) and reshape to \((H',W',C')\), producing \(\mathbf{z}_{\mathbf{q}_t}\). The scalar value \(v_t\) is embedded via broadcast: we set every element of a latent frame to the same normalized value, yielding \(\mathbf{z}_{v_t}\).

\paragraph{Latent sequence during training.}
During training, we assemble a fixed-length sequence of latent frames that includes both conditioning and target frames. Let \(K\) denote a fixed prediction horizon. The sequence is:
\[
[\,\mathbf{z}_{\text{blank}},\; \mathbf{z}_{\mathbf{q}_t},\; \mathbf{z}_{\mathbf{o}_t^1},\; \mathbf{z}_{\mathbf{o}_t^2},\; \mathbf{z}_{\mathbf{o}_t^3},\; \mathbf{z}_{\mathbf{q}_{t+K}},\; \mathbf{z}_{v_t}\,],
\]
where \(\mathbf{z}_{\text{blank}}\) is a zero-initialized placeholder required by the causal VAE. The first five frames (blank, current proprioception, and current images) serve as clean conditioning, while the remaining two frames (future proprioception \(\mathbf{z}_{\mathbf{q}_{t+K}}\) and value \(\mathbf{z}_{v_t}\)) are corrupted with Gaussian noise at a randomly sampled level \(\sigma\). The denoiser \(D_{\theta}\) learns to recover the clean targets from the noisy ones, conditioned on the clean prefix.


\paragraph{Latent sequence during inference.}
At inference time, only the conditioning frames are available. We encode the current observations (images and proprioception) into their respective latent frames, form the same prefix $[\mathbf{z}_{\text{blank}}, \mathbf{z}_{\mathbf{q}_t}, \mathbf{z}_{\mathbf{o}_t^1}, \allowbreak \mathbf{z}_{\mathbf{o}_t^2}, \mathbf{z}_{\mathbf{o}_t^3}]$, and run reverse diffusion to generate the target frames $\hat{\mathbf{z}}_{\mathbf{q}_{t+K}}$ and $\hat{\mathbf{z}}_{v_t}$. The predicted value $\hat{v}_t$ is obtained by averaging all elements of $\hat{\mathbf{z}}_{v_t}$ and rescaling from $[-1,1]$ back to $[0,1]$. To recover the future proprioceptive state $\hat{\mathbf{q}}_{t+K}$, we apply the inverse of the repeat-padding injection: flatten $\hat{\mathbf{z}}_{\mathbf{q}_{t+K}}$, split into consecutive chunks of size equal to the original proprioception dimension, average each chunk, and rescale to the original range.

\paragraph{Training objective.}
We adopt the flow matching formulation as in Wan2.2~\citep{wan2025wan}.
Let $\mathbf{z}_0$ denote a clean latent frame (either
$\mathbf{z}_{\mathbf{q}_{t+K}}$ or $\mathbf{z}_{v_t}$), and let $\mathbf{z}_1 \sim \mathcal{N}(\mathbf{0}, \mathbf{I})$
be a Gaussian noise latent of the same shape.
We construct a linear interpolation path
\[
\mathbf{z}_\tau = (1-\tau)\mathbf{z}_0 + \tau \mathbf{z}_1,
\quad \tau \in [0,1].
\]
The model $v_\theta(\mathbf{z}_\tau; \tau, \mathbf{c})$
is trained to predict the constant velocity
$\mathbf{z}_1 - \mathbf{z}_0$ along this path.
The overall objective is a weighted combination:
\[
\begin{aligned}
\mathcal{L} = &\;
\lambda_{\text{prop}}
\mathbb{E}_{\mathbf{z}_0^{\mathbf{q}}\sim p_{\text{data}},
\mathbf{z}_1\sim\mathcal N(\mathbf{0},\mathbf I),
\tau\sim\mathcal U[0,1]}
\left[
\| v_\theta(\mathbf{z}_\tau^{\mathbf{q}}; \tau, \mathbf{c})
- (\mathbf{z}_1 - \mathbf{z}_0^{\mathbf{q}}) \|_2^2
\right] \\
&+
\lambda_{\text{val}}
\mathbb{E}_{\mathbf{z}_0^{v}\sim p_{\text{data}},
\mathbf{z}_1\sim\mathcal N(\mathbf{0},\mathbf I),
\tau\sim\mathcal U[0,1]}
\left[
\| v_\theta(\mathbf{z}_\tau^{v}; \tau, \mathbf{c})
- (\mathbf{z}_1 - \mathbf{z}_0^{v}) \|_2^2
\right].
\end{aligned}
\]
where $\mathbf{z}_\tau^{\mathbf{q}}$ and $\mathbf{z}_\tau^{v}$
are modality-specific interpolated latents,
$\mathbf{c}$ denotes the clean conditioning frames,
and $\tau \sim \mathcal{U}[0,1]$ is the flow time step.
We also experimented with jointly predicting future visual latents, but observed a degradation in value estimation accuracy. We hypothesize that this is due to the inherent difficulty mismatch between the two tasks: visual generation requires capturing high-dimensional spatial structure, while the value latent has a much simpler structure and is more susceptible to interference from the visual reconstruction objective during joint optimization.

By treating all modalities as latent frames, our architecture repurposes a powerful video generator for value estimation while preserving its spatiotemporal priors. The inclusion of future proprioceptive prediction serves two purposes: it forces the model to internalize the robot's own dynamics, which is essential for tasks requiring precise limb coordination, and it provides an implicit measure of motion that complements visual cues for value estimation. 

\subsection{Reward Definition and Value Training}

We now define the learning targets for our video-generative value model. Each episode in the training data is annotated with a binary success label indicating the final task outcome. For an episode of length $T$, we define the step-wise reward $r_t$ to encode both temporal progress and completion status:
\begin{equation}
r_t =
\begin{cases}
\frac{1}{T} , & \text{if } t < T, \\[4pt]
0, & \text{if } t = T \text{ and success}, \\[4pt]
1, & \text{if } t = T \text{ and failure},
\end{cases}
\end{equation}
where $t = 1, \dots, T$. Under this formulation, the cumulative return $G_t = \sum_{k=t}^{T} r_k$ provides a discriminative supervision signal that distinguishes outcomes through distinct value ranges:
\begin{equation}
G_t =
\begin{cases}
\frac{T-t}{T}, & \text{if success}, \\[4pt]
\frac{T-t}{T} + 1, & \text{if failure}.
\end{cases}
\end{equation}


Under this formulation, $G_t$ reflects normalized task progress within $[0, 1)$ for successful episodes, while failed episodes are shifted to $[1, 2)$ by the terminal penalty. This ensures a constant margin of $1.0$ between outcomes at any temporal stage, effectively resolving the ambiguity between progress and failure in value estimation.

The return $G_t$ serves as the supervision signal for the value latent $\mathbf{z}_{v_t}$, which is treated as the clean target in the flow matching objective described in Sec.~\ref{sec:overall_arch}. This formulation provides a consistent and outcome-aware supervision signal across episodes of varying lengths. By jointly predicting the return and future proprioception, the model learns to capture both task-level integrity and the robot's embodied dynamics, effectively grounding value estimation in anticipated embodiment evolution.

\section{Experiments}

\subsection{Experimental Setup}

\subsubsection{Evaluation Tasks}
\label{sec:eval_tasks}
\begin{figure}[t]
    \centering
    \includegraphics[width=1.0\linewidth]{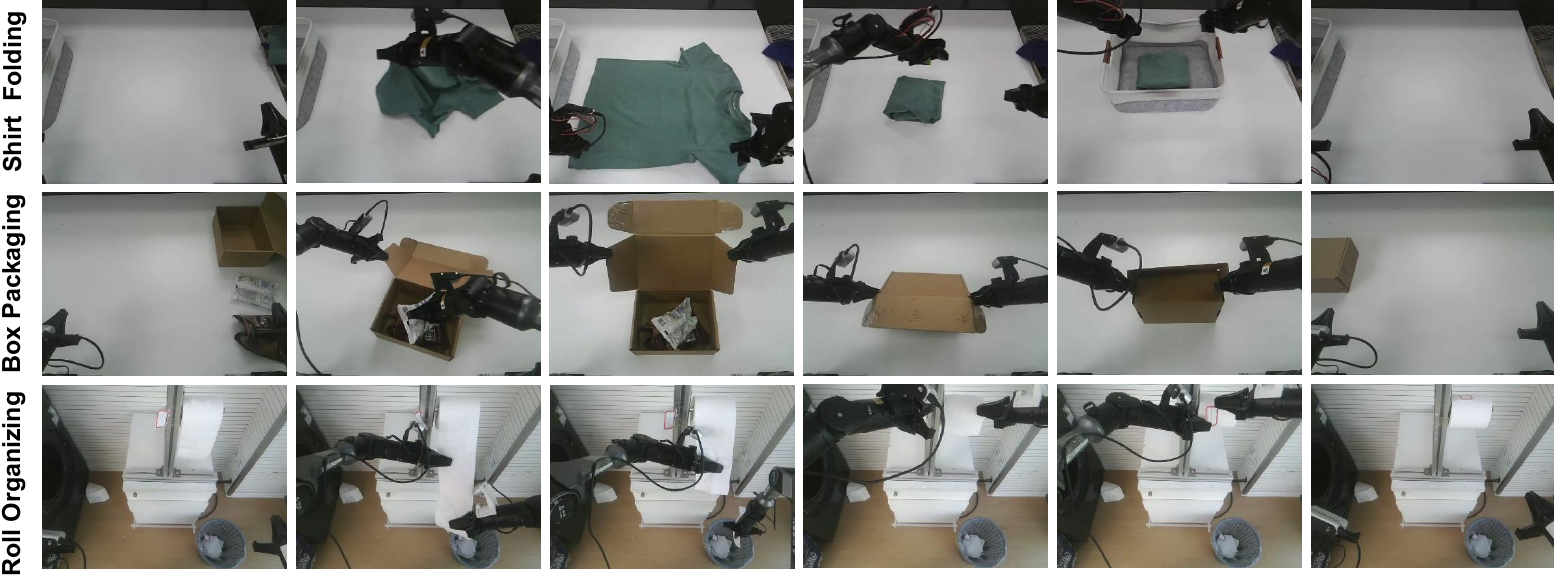}
    \caption{Illustration of the three real-world tasks. For each task, we show the initial state (left), key intermediate stages (middle), and the final successful state (right). }
    \label{fig:tasks}
\end{figure}

Our evaluations and real-robot experiments span three long-horizon manipulation tasks: shirt folding, box packaging, and paper roll organization, summarized below and illustrated in Figure~\ref{fig:tasks}.

\paragraph{Shirt folding.} This task evaluates dual-arm coordination for manipulating highly deformable textiles. The robot must first take a garment from a basket, flatten it on the table, execute a folding sequence (sleeves and sides inward, then longitudinal and cross folds), and return the folded shirt to the basket. Success requires neat folding and placement in the basket within 300 seconds. Failure occurs if entanglement damages the garment, fold collapses, or return is improper.

\paragraph{Box packaging.} This task evaluates long-horizon dual-arm coordination through a multi-stage manipulation sequence. The robot picks a target item, places it into a cardboard box, then folds the side flaps and closes the lid. Success requires the item to be fully enclosed in a structurally sound box with all tabs interlocked within 300 seconds. Failure occurs if the item is dropped, the box is damaged, or the box cannot be fully sealed.

\paragraph{Paper roll organization.} This task evaluates precise, multi-stage manipulation of flexible paper. The robot must grasp and tear off a single sheet, discarding it into a receptacle, then collaboratively rewind the remaining loose end until it is flush with the roll. Finally, a sealing sticker is applied to secure the end. Success requires completing the tear, disposal, and sealing within 240 seconds. Excessive tearing or failure to secure the sticker is recorded as a failure.

\subsubsection{Evaluation Dataset}
\label{sec:datasets}

\begin{figure*}[t]
    \centering
    \includegraphics[width=0.88\textwidth]{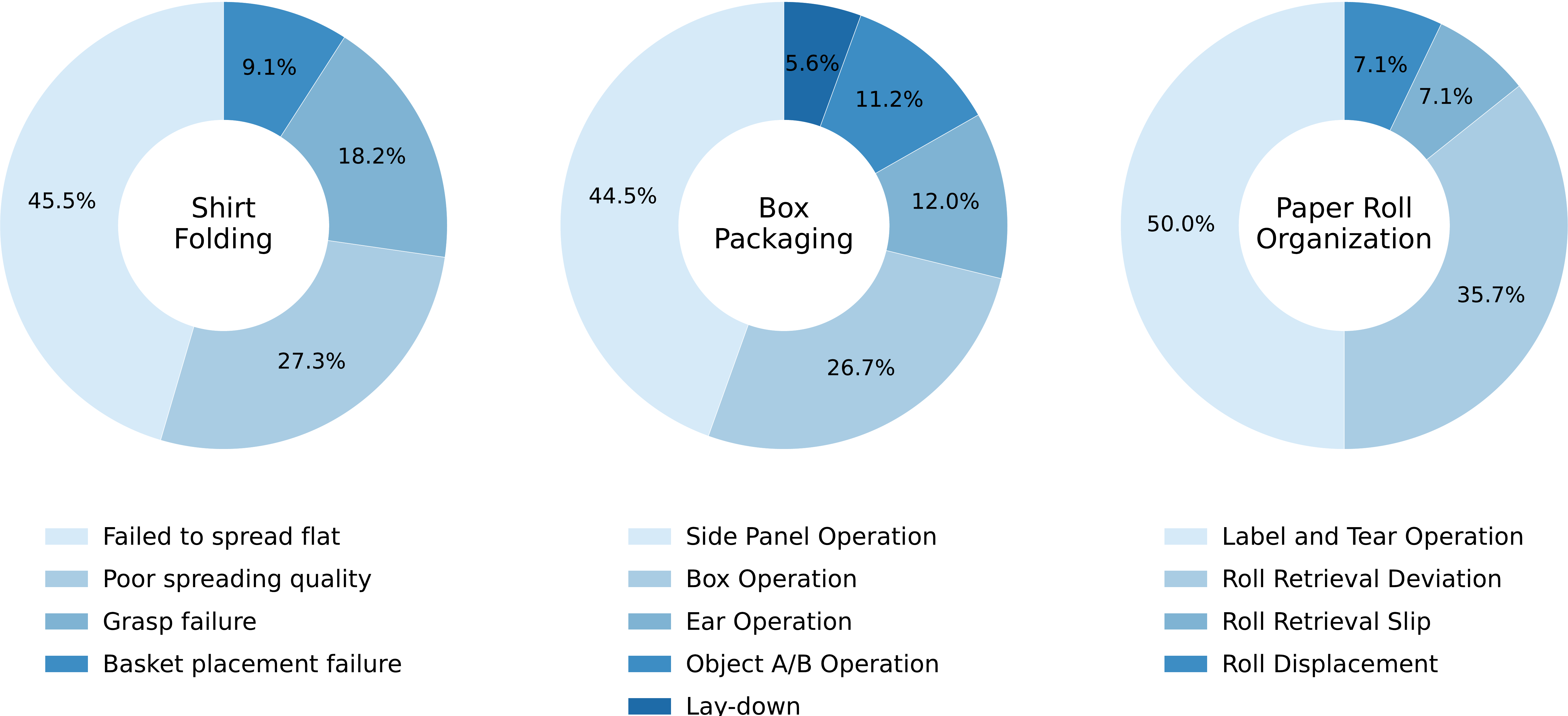}
    \caption{Error type distribution for shirt folding, box packaging, and paper roll organization. Each pie chart shows the relative frequency of failure modes, based on 50 held-out episodes per task.}
    \label{fig:error_distribution}
\end{figure*}

We use 50 held-out episodes per task for evaluation. Milestone and error frames are pre-defined per task and manually annotated for each episode. Milestone frames mark key task stages and appear in every episode; error frames, by contrast, are annotated only for episodes in which the corresponding failure mode actually occurs. Figure~\ref{fig:error_distribution} reports the error type distribution for each task.

\paragraph{Shirt folding.} Failures are concentrated in the flattening stage, with Failed to spread flat (45.5\%) and Poor spreading quality (27.3\%) being the most frequent, followed by Grasp failure and Basket placement failure. The key challenge lies in reliably flattening a highly deformable textile without residual wrinkles or misalignment.

\paragraph{Box packaging.} Failures span multiple sub-tasks, dominated by Side panel operation (44.5\%) and Box operation (26.7\%), with additional errors in Ear operation, Object A/B operation, and Lay-down. The difficulty stems from the need to coordinate a sequential, multi‑stage assembly where each step must be executed precisely before the next can begin.

\paragraph{Paper roll organization.} The majority of failures involve tearing and retrieval: Label and tear operation (50.0\%) and Roll retrieval deviation (35.7\%), along with Roll retrieval slip and Roll displacement. The core challenge is fine‑grained manipulation of flexible paper, where even small deviations in tearing angle or retrieval trajectory can ruin the product.

\subsubsection{Implementation Details}
\label{sec:impl_details}

For the VLM-based value model, we follow the same design as $\pi_{0.6}^*$~\citep{intelligence2025pi0.6}, formulating value estimation as a 201-way classification problem over return bins. Both this baseline and our ViVa-based variant are trained within the identical RECAP pipeline. We conduct two rounds of rollouts: the first round collects 150 successful and 50 failed episodes per task, and the second round collects 100 successful and 50 failed episodes per task. The rollout data are combined with demonstration data from all three tasks. All models are trained for one epoch with batch size 192. The prediction horizon $K$ is set to 50 for all tasks, aligning with the default horizon used for advantage estimation in the RECAP framework. For ViVa, the loss weights for future proprioception and value prediction are set to $\lambda_{\text{prop}}=0.5$ and $\lambda_{\text{val}}=1.0$, respectively. During inference, we use 1 denoising step for ViVa with DDIM sampling. All experiments are conducted on 8 NVIDIA A800 GPUs.

\subsection{Evaluation Metrics}

\paragraph{Event Response Score (ERS).}
For each evaluation episode, we annotate a set of key frames $\mathcal{K}$, each labeled with an expected direction $d_k \in \{+1, -1\}$: $d_k = +1$ for milestones and $d_k = -1$ for execution errors.
Let $V_{\text{before}}^{(k)}$ and $V_{\text{after}}^{(k)}$ denote the mean predicted value over a window of $w$ frames immediately preceding and following the annotated frame.
The ERS for key frame $k$ is defined as
\begin{equation}
\mathrm{ERS}_k = d_k \cdot \frac{V_{\text{after}}^{(k)} - V_{\text{before}}^{(k)}}{V_{\text{before}}^{(k)} + \epsilon},
\end{equation}
where $\epsilon$ is a small positive constant to avoid division by zero.
We then define \textit{Milestone Sensitivity (MS)} as the average ERS over all milestone frames ($d_k = +1$), and \textit{Error Sensitivity (ES)} as the average ERS over all error frames ($d_k = -1$).
Positive values indicate correct directional responses, with larger magnitudes reflecting stronger sensitivity; negative values indicate incorrect ones.

\paragraph{Event Detection Rate (EDR).} Based on the ERS, we consider an annotated event frame $k$ as detected if $\mathrm{ERS}_k > 0$. Using this binary detection criterion, we define the \textit{Milestone Detection Rate (MDR)} and \textit{Error Detection Rate (ErrDR)} as the proportions of correctly detected frames within each event type:
\begin{equation}
\mathrm{MDR} = \frac{N_{\mathrm{milestone}}^{+}}{N_{\mathrm{milestone}}},
\qquad
\mathrm{ErrDR} = \frac{N_{\mathrm{error}}^{+}}{N_{\mathrm{error}}},
\end{equation}
where $N_{\mathrm{milestone}}^{+}$ and $N_{\mathrm{error}}^{+}$ denote the numbers of milestone frames ($d_k = +1$) and error frames ($d_k = -1$) with positive ERS values, respectively, and $N_{\mathrm{milestone}}$, $N_{\mathrm{error}}$ are the total counts of milestone and error frames.

\subsection{Value Model Evaluation}
\paragraph{Evaluation setup.}
We benchmark ViVa against recent state-of-the-art value models on three tasks: shirt folding, box packaging, and paper organization. 

\Needspace{8\baselineskip}
\begin{wraptable}{r}{0.55\textwidth}
\vspace{-12pt}
\tablestyle{2.5pt}{1.2}
\caption{Comparison with robotics SOTA value models. Sparse‑sampled methods show low or negative correlation. Dense‑supervised approaches perform much better, and ViVa establishes new SOTA on two of three tasks.}
\label{tab:sota_value_model}
\begin{tabular}{ccccccc}
\hline
 & \multicolumn{2}{c}{Shirt} & \multicolumn{2}{c}{Box} & \multicolumn{2}{c}{Paper} \\ \cline{2-7}
\multirow{-2}{*}{Method} & Pearson & Spearman & Pearson & Spearman & Pearson & Spearman \\ \hline
\multicolumn{7}{c}{Sparse-sampled} \\ \hline
GVL & 0.276 & 0.263 & -0.081 & -0.114 & -0.099 & -0.096 \\
TopReward & 0.118 & 0.090 & 0.503 & 0.499 & 0.318 & 0.241 \\ \hline
\multicolumn{7}{c}{Dense-supervised} \\ \hline
$\pi^{*}_{0.6}$ & 0.740 & 0.720 & \textbf{0.946} & \textbf{0.945} & 0.926 & 0.936 \\
\rowcolor[HTML]{EFF6FB}
ViVa & \textbf{0.984} & \textbf{0.982} & 0.868 & 0.850 & \textbf{0.952} & \textbf{0.948} \\ \hline
\end{tabular}
\vspace{-10pt}
\end{wraptable}
GVL~\citep{ma2025vision} and TopReward~\citep{chen2026topreward} leverage VLMs for sparse-sampled value estimation, while $\pi^{*}_{0.6}$~\citep{intelligence2025pi0.6} and ViVa are trained on $\sim$20 hours of mixed demonstration data from all three tasks. We use 50 held-out episodes per task and manually annotate milestone and error frames for evaluation. See Sec.~\ref{sec:datasets} for dataset details.

\paragraph{Quantitative comparison.}
Table~\ref{tab:sota_value_model} reports Pearson and Spearman correlation as a global measure of value quality. GVL and TopReward yield low or even negative correlations, while $\pi^{*}_{0.6}$ and ViVa achieve consistently strong correlations, with ViVa leading on two of three tasks and $\pi^{*}_{0.6}$ ahead on box packaging. This performance gap stems from GVL and TopReward being training-free methods that query VLMs at sparse intervals, lacking the dense temporal supervision needed to reliably associate static frames with task progress. In contrast, dense supervision on every frame enables $\pi^{*}_{0.6}$ and ViVa to capture the continuous evolution of manipulation behaviors.

\begin{figure}[t]
    \centering
    \includegraphics[width=1.0\linewidth]{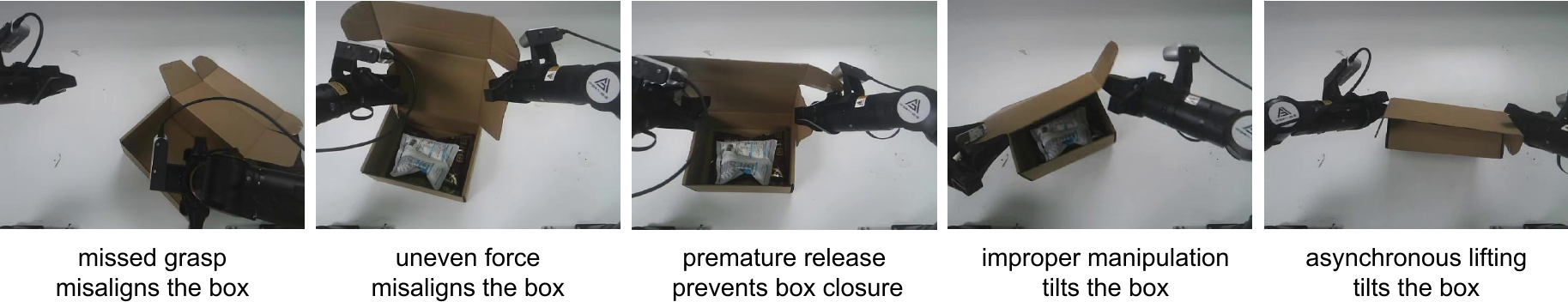}
    \caption{Detection of subtle manipulation errors by ViVa. The model produces clear value drops at failure moments, confirming sensitivity to fine-grained execution errors.}
    \label{fig:error_detection}
\end{figure}

For $\pi^{*}_{0.6}$ and ViVa, however, higher global correlation does not necessarily imply better value quality. Near-uniform predictions collapse advantages to zero, providing no RL signal. We therefore evaluate on four fine-grained event-level metrics for a more faithful comparison. Tables~\ref{tab:event_sensitivity} and~\ref{tab:event_detection} show ViVa outperforming $\pi^{*}_{0.6}$ across nearly all metrics. On shirt folding, $\pi^{*}_{0.6}$ yields negative MS ($-0.266$) and ES ($-0.299$), indicating that its predictions move in the wrong direction at key events, while ViVa achieves positive values on both. For detection, ViVa attains MDR of 0.990 and ErrDR of 1.000, nearly double $\pi^{*}_{0.6}$'s 0.407 and 0.500. This gap persists across tasks, and Fig.~\ref{fig:error_detection} qualitatively confirms ViVa's sensitivity to subtle execution errors.


\begin{table}[h]
\vspace{5pt}
\centering
\begin{minipage}{0.485\textwidth}
\tablestyle{2.3pt}{1.2}
\caption{Event-sensitivity comparison. MS (milestone sensitivity) and ES (error sensitivity) measure directional response quality. ViVa achieves substantially higher sensitivity than the VLM baseline across all three tasks.}
\label{tab:event_sensitivity}
\begin{tabular}{cccccccc}
\hline
 &  & \multicolumn{2}{c}{Shirt} & \multicolumn{2}{c}{Box} & \multicolumn{2}{c}{Paper} \\ \cline{3-8}
\multirow{-2}{*}{Method} & \multirow{-2}{*}{Backbone} & MS & ES & MS & ES & MS & ES \\ \hline
$\pi^{*}_{0.6}$ & VLM & -0.266 & -0.299 & 0.013 & 0.006 & 0.023 & 0.028 \\
\rowcolor[HTML]{EFF6FB}
ViVa & Video-gen & \textbf{0.095} & \textbf{0.012} & \textbf{0.034} & \textbf{0.026} & \textbf{0.039} & \textbf{0.055} \\ \hline
\end{tabular}
\end{minipage}
\hfill
\begin{minipage}{0.485\textwidth}
\tablestyle{2.7pt}{1.2}
\caption{Event-detection comparison. MDR (milestone detection rate) and ErrDR (error detection rate) measure the proportion of correctly detected event frames. ViVa consistently achieves higher detection rates across all tasks.}
\label{tab:event_detection}
\begin{tabular}{cccccccc}
\hline
 &  & \multicolumn{2}{c}{Shirt} & \multicolumn{2}{c}{Box} & \multicolumn{2}{c}{Paper} \\ \cline{3-8}
\multirow{-2}{*}{Method} & \multirow{-2}{*}{Backbone} & MDR & ErrDR & MDR & ErrDR & MDR & ErrDR \\ \hline
$\pi^{*}_{0.6}$ & VLM & 0.407 & 0.500 & \textbf{0.765} & 0.506 & 0.704 & 0.667 \\
\rowcolor[HTML]{EFF6FB}
ViVa & Video-gen & \textbf{0.990} & \textbf{1.000} & 0.733 & \textbf{0.612} & \textbf{0.762} & \textbf{0.778} \\ \hline
\end{tabular}
\end{minipage}
\vspace{5pt}
\end{table}

\begin{figure}[b]
    \centering
      \vspace{-8pt}
    \includegraphics[width=1.0\linewidth]{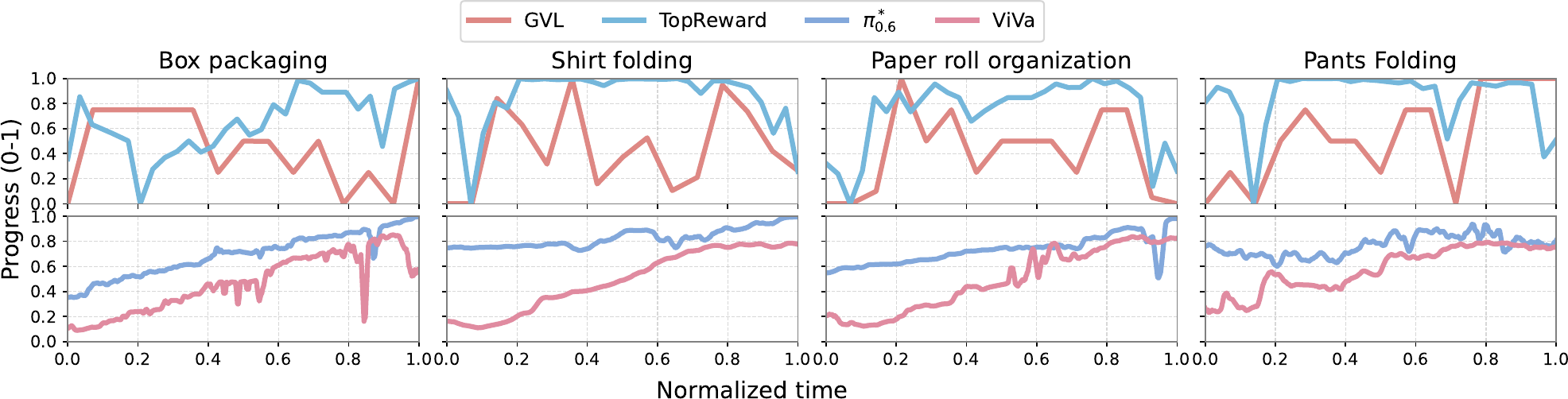}
    \caption{Value trajectory visualization across all tasks. Trajectories for $\pi^{*}_{0.6}$ and ViVa are smoothed by 20\%. ViVa tracks task progress with clear responses at key events and maintains sensitivity on the unseen task, demonstrating stronger generalization.}
    \label{fig:trajectory_comparison}
     \vspace{-8pt}
\end{figure}

\paragraph{Trajectory visualization.}
Figure~\ref{fig:trajectory_comparison} visualizes the predicted value trajectories of all four methods on the three in-domain tasks.
GVL and TopReward, which query VLMs at sparse intervals, produce noisy and flat value curves that fail to reflect task progression.
$\pi^{*}_{0.6}$ captures the overall trend with dense supervision, but its responses at fine-grained events are often delayed or muted.
In contrast, ViVa is more sensitive to fine-grained manipulation events, with clear value increases at milestones and decreases at execution errors. This suggests that grounding value in anticipated dynamics contributes to more precise progress awareness.

\subsection{Object Generalization}
\paragraph{Evaluation setup.}
We further evaluate $\pi^{*}_{0.6}$ and ViVa on an out-of-domain pants folding task comprising 50 held-out episodes annotated with milestone frames. We focus on milestones, as error detection on novel objects remains challenging without large-scale pretraining.

\begin{wraptable}{r}{0.485\textwidth}
\tablestyle{10pt}{1.2}
\caption{Object generalization on the out-of-domain pants folding task. ViVa consistently outperforms $\pi^{*}_{0.6}$ across all evaluation metrics.}
\label{tab:obj_generalization}
\begin{tabular}{ccccc}
\hline
Method & Pearson & Spearman & MS & MDR \\ \hline
$\pi^{*}_{0.6}$ & 0.740 & 0.721 & 0.016 & 0.397 \\
\rowcolor[HTML]{EFF6FB}
ViVa & \textbf{0.819} & \textbf{0.799} & \textbf{0.037} & \textbf{0.643} \\ \hline
\end{tabular}
\vspace{-10pt}
\end{wraptable}

\paragraph{Quantitative comparison.}
Table~\ref{tab:obj_generalization} reports object generalization results on the out-of-domain pants folding task.
ViVa outperforms $\pi^{*}_{0.6}$ across all four metrics, with the largest gains on fine-grained event metrics: MS more than doubles (0.037 vs.\ 0.016) and MDR improves by over 60\% (0.643 vs.\ 0.397) on this out-of-domain task.

\paragraph{Trajectory visualization.}
The pants folding panel in Figure~\ref{fig:trajectory_comparison} shows value trajectories on the out-of-domain task.
GVL and TopReward remain unresponsive, consistent with their general inability to capture temporal progress.
$\pi^{*}_{0.6}$ captures the overall trend but misses key milestones.
In contrast, ViVa consistently responds to milestone events, demonstrating that its spatiotemporal priors provide a more transferable representation for generalization.

\subsection{Real-Robot Experiments}
\paragraph{Setup.}
We evaluate on three real-world tasks: shirt folding, box packaging, and paper roll organization. We compare imitation learning baselines Gigabrain-0~\citep{team2025gigabrain} and $\pi_{0.5}$~\citep{intelligence2025pi} against two RECAP~\citep{intelligence2025pi0.6} variants, both built on Gigabrain-0 with different value models: a VLM-based value function versus ViVa. Both RECAP variants undergo two rounds of rollouts. See Sec.~\ref{sec:impl_details}.

\begin{table}[h]
\vspace{5pt}
\centering
\begin{minipage}{0.485\textwidth}
\tablestyle{5.3pt}{1.2}
\caption{Real-robot experiment results. Success rates (\%) on three manipulation tasks. RECAP (ViVa) outperforms both imitation learning baselines and RECAP (VLM) on all tasks.}
\label{tab:real_robot}
\begin{tabular}{ccccc}
\hline
Method & Shirt (\%) & Box (\%) & Paper (\%) & Avg (\%) \\ \hline
\multicolumn{5}{c}{Imitation learning} \\ \hline
$\pi_{0.5}$ & 60.0 & 40.0 & 40.0 &  46.7 \\
Gigabrain-0 & 50.0 & 50.0 & 20.0 & 40.0 \\ \hline
\multicolumn{5}{c}{Reinforcement Learning} \\ \hline
RECAP (VLM) & 70.0 & 60.0 & 60.0 & 63.3 \\
\rowcolor[HTML]{EFF6FB}
RECAP (ViVa) & \textbf{90.0} & \textbf{70.0}  & \textbf{80.0}  & \textbf{80.0} \\ \hline
\end{tabular}
\end{minipage}
\hfill
\begin{minipage}{0.485\textwidth}
\tablestyle{1.5pt}{1.2}
\caption{Ablation on architecture design for box packaging. Replacing the VLM backbone with VG alone improves milestone metrics, and adding future state prediction further boosts error sensitivity and detection. VG denotes video generator.}
\label{tab:arch_box}
\begin{tabular}{cccccccc}
\hline
\shortstack{Method\\\mbox{}} & \shortstack{Backbone\\\mbox{}} & \shortstack{\strut Input\\State} & \shortstack{\strut Future\\State} & \shortstack{MS\\\mbox{}} & \shortstack{ES\\\mbox{}} & \shortstack{MDR\\\mbox{}} & \shortstack{ErrDR\\\mbox{}} \\ \hline
$\pi^{*}_{0.6}$ & VLM & \ding{55} & \ding{55} & 0.013 & 0.006 & 0.765 & 0.506 \\
VG-only & VG & \ding{55} & \ding{55} & \textbf{0.048} & 0.004 & \textbf{0.853} & 0.449 \\
ViVa w/o fut & VG & \ding{51} & \ding{55} & 0.047 & 0.008 & 0.778 & 0.527 \\
\rowcolor[HTML]{EFF6FB}
ViVa & VG & \ding{51} & \ding{51} & 0.034 & \textbf{0.026} & 0.733 & \textbf{0.612} \\ \hline
\end{tabular}
\end{minipage}
\end{table}

\paragraph{Main results.}
Table~\ref{tab:real_robot} reports the success rates on three long-horizon manipulation tasks. Both RECAP variants substantially outperform the imitation learning baselines, confirming the benefit of reinforcement learning. Replacing the VLM-based value function with ViVa yields further gains across all tasks, demonstrating that better value estimation translates to stronger policy improvement.

\begin{figure*}[b]
    \centering
    \setlength{\tabcolsep}{2pt}

    \begin{tabular}{ccc}
        \includegraphics[width=0.32\textwidth]{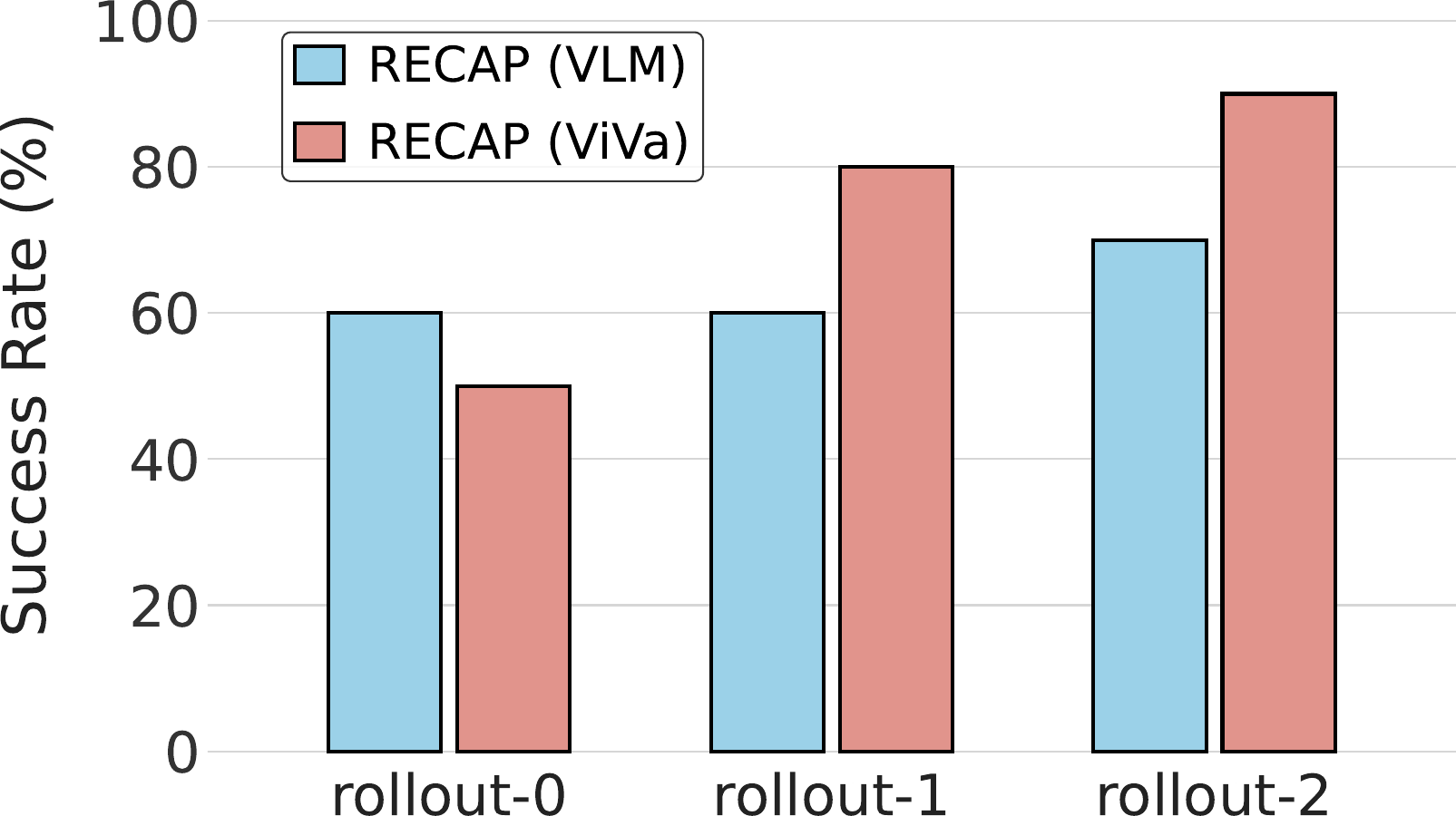} &
        \includegraphics[width=0.32\textwidth]{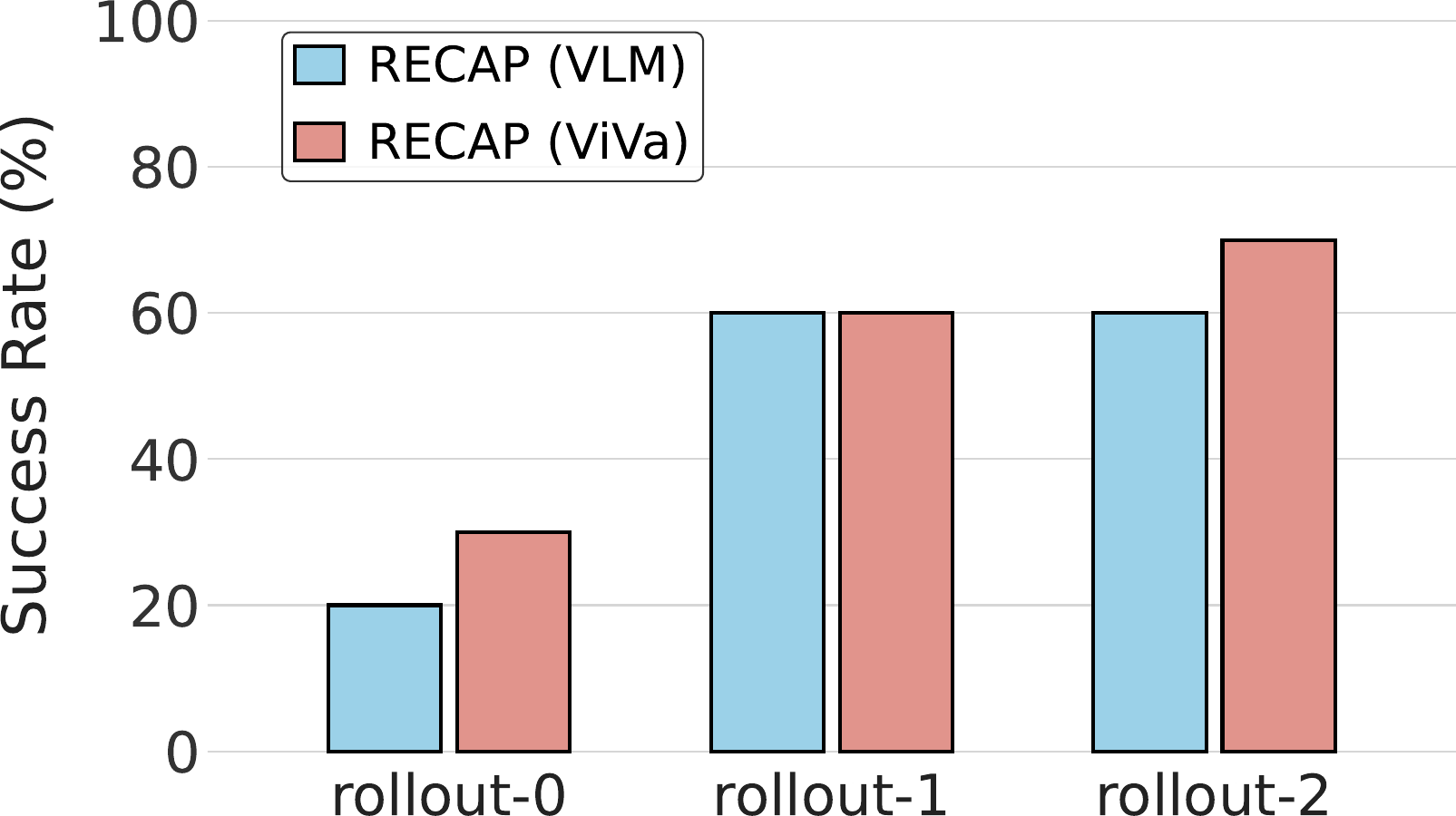} &
        \includegraphics[width=0.32\textwidth]{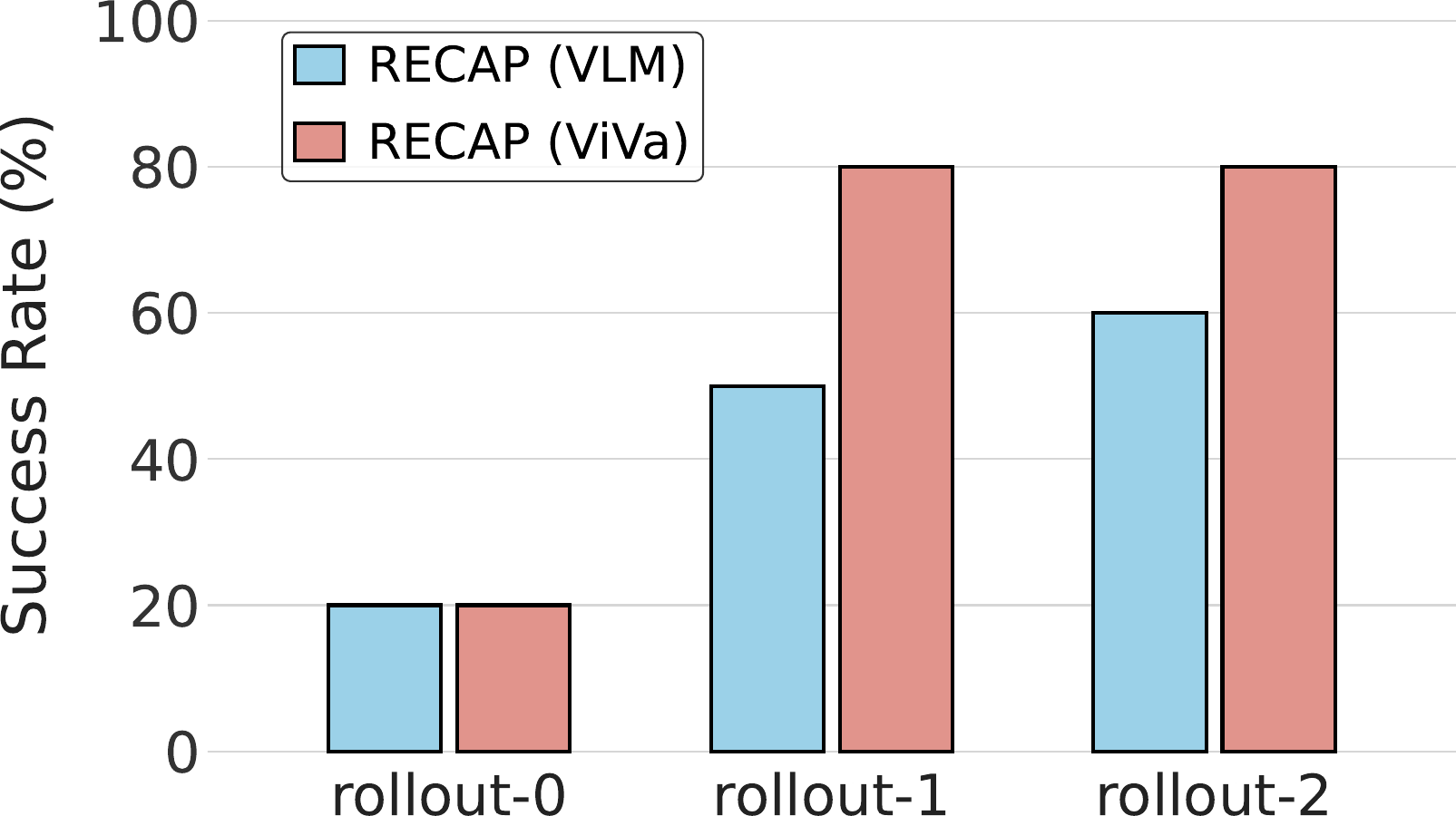} \\
        {\small (a) Shirt folding.} &
        {\small (b) Box packaging.} &
        {\small (c) Paper roll organization.}
    \end{tabular}
    \caption{Rollout-wise performance evolution across three manipulation tasks. Both methods improve over successive rounds, with RECAP (ViVa) achieving the best overall performance.}
    \label{fig:rollout_evolution}
\end{figure*}

\paragraph{Rollout-wise performance evolution.}
Figure~\ref{fig:rollout_evolution} shows the success rates across rollout rounds. ViVa improves substantially faster from R0 to R1: on shirt folding, it jumps from 50\% to 80\% while VLM stagnates at 60\%; on paper roll organization, it surges from 20\% to 80\% vs.\ VLM's 20\% to 50\%. From R1 to R2, VLM plateaus while ViVa keeps improving, reaching 90\% on shirt and extending its lead on box. Overall, ViVa converges faster to stronger final performance.

\subsection{Ablation Studies}
\paragraph{Effect of video generator backbone.}
We conduct architecture ablation on all three tasks, as reported in Tables~\ref{tab:arch_box}, \ref{tab:arch_shirt} and~\ref{tab:arch_paper}. Across the board, replacing the VLM backbone with a video generator brings consistent gains. On box packaging, the video backbone alone lifts MS from 0.013 to 0.048 and MDR from 0.765 to 0.853. The improvement is even more pronounced on shirt folding, where the VLM baseline suffers from negative MS ($-0.266$) and ES ($-0.299$), meaning its predictions move in the wrong direction at key events. Switching to a video generator reverses this to a positive MS of 0.137 and lifts MDR from 0.407 to 1.000. On paper roll organization, the video backbone alone yields mixed results, with the full architecture providing the clearest gains (discussed next). These results confirm that video generative priors offer a substantially stronger foundation for value estimation than static VLM representations.

\begin{table}[t]
\centering
\begin{minipage}{0.485\textwidth}
\tablestyle{1.5pt}{1.2}
\caption{Ablation on architecture design for shirt folding. VG-only outperforms the VLM baseline across all metrics, with further gains from future state prediction. VG denotes video generator.}
\label{tab:arch_shirt}
\begin{tabular}{cccccccc}
\hline
\shortstack{Method\\\mbox{}} & \shortstack{Backbone\\\mbox{}} & \shortstack{\strut Input\\State} & \shortstack{\strut Future\\State} & \shortstack{MS\\\mbox{}} & \shortstack{ES\\\mbox{}} & \shortstack{MDR\\\mbox{}} & \shortstack{ErrDR\\\mbox{}} \\ \hline
$\pi^{*}_{0.6}$ & VLM & \ding{55} & \ding{55} & -0.266 & -0.299 & 0.407 & 0.500 \\
VG-only & VG & \ding{55} & \ding{55} & \textbf{0.137} & -0.026 & \textbf{1.000} & 0.500 \\
ViVa w/o fut & VG & \ding{51} & \ding{55} & 0.093 & 0.006 & 0.998 & \textbf{1.000} \\
\rowcolor[HTML]{EFF6FB}
ViVa & VG & \ding{51} & \ding{51} & 0.095 & \textbf{0.012} & 0.990 & \textbf{1.000} \\ \hline
\end{tabular}
\end{minipage}
\hfill
\begin{minipage}{0.485\textwidth}
\tablestyle{1.7pt}{1.2}
\caption{Ablation on architecture design for paper roll organization. ViVa with full architecture achieves the best performance across all metrics. VG denotes video generator.}
\label{tab:arch_paper}
\begin{tabular}{cccccccc}
\hline
\shortstack{Method\\\mbox{}} & \shortstack{Backbone\\\mbox{}} & \shortstack{\strut Input\\State} & \shortstack{\strut Future\\State} & \shortstack{MS\\\mbox{}} & \shortstack{ES\\\mbox{}} & \shortstack{MDR\\\mbox{}} & \shortstack{ErrDR\\\mbox{}} \\ \hline
$\pi^{*}_{0.6}$ & VLM & \ding{55} & \ding{55} & 0.023 & 0.028 & 0.704 & 0.667 \\
VG-only & VG & \ding{55} & \ding{55} & 0.016 & -0.011 & 0.651 & 0.444 \\
ViVa w/o fut & VG & \ding{51} & \ding{55} & 0.019 & -0.043 & 0.644 & 0.444 \\
\rowcolor[HTML]{EFF6FB}
ViVa & VG & \ding{51} & \ding{51} & \textbf{0.039} & \textbf{0.055} & \textbf{0.762} & \textbf{0.778} \\ \hline
\end{tabular}
\end{minipage}
\end{table}

\begin{wraptable}{r}{0.485\textwidth}
\vspace{-12pt}
\tablestyle{3.7pt}{1.2}
\caption{Ablation on pretrained weights for box packaging. Removing pretrained weights degrades performance across all metrics, confirming the importance of spatiotemporal priors.}
\label{tab:pretrain_box}
\begin{tabular}{cccccc}
\hline
Method & Pretrained & MS & ES & MDR & ErrDR \\ \hline
ViVa w/o pretrain & \ding{55} & 0.009 & 0.023 & 0.523 & 0.524 \\
\rowcolor[HTML]{EFF6FB}
ViVa & \ding{51} & \textbf{0.034} & \textbf{0.026} & \textbf{0.733} & \textbf{0.612} \\ \hline
\end{tabular}
\end{wraptable}

\paragraph{Effect of input and future state.}
Building on the video generator backbone, we further examine the effect of incorporating proprioceptive state and future dynamics prediction across all three tasks (Tables~\ref{tab:arch_box}, \ref{tab:arch_shirt}, \ref{tab:arch_paper}). On box packaging, adding input state lifts ErrDR from 0.449 to 0.527, and future state prediction further raises ES to 0.026 and ErrDR to 0.612. On shirt folding, the full architecture brings ES from $-0.026$ (VG-only) to $0.012$, turning a negative error sensitivity into a positive one, while maintaining near-perfect MDR (0.990) and ErrDR (1.000). On paper roll organization, the full ViVa achieves the strongest results across all metrics (MS 0.039, ES 0.055, MDR 0.762, ErrDR 0.778), while the VG-only variant yields negative ES ($-0.011$). These consistent improvements confirm that future dynamics prediction sharpens error awareness beyond what the video backbone alone provides, making the value model more sensitive to task-relevant execution events.

\begin{table}[b]
\centering
\begin{minipage}{0.485\textwidth}
\tablestyle{3.8pt}{1.2}
\caption{Ablation on pretrained weights for shirt folding. Removing pretrained weights degrades performance, showing spatiotemporal priors' value.}
\label{tab:pretrain_shirt}
\begin{tabular}{cccccc}
\hline
Method & Pretrained & MS & ES & MDR & ErrDR \\ \hline
ViVa w/o pretrain & \ding{55} & 0.117 & -0.524 & 0.770 & 0.500 \\
\rowcolor[HTML]{EFF6FB}
ViVa & \ding{51} & \textbf{0.095} & \textbf{0.012} & \textbf{0.990} & \textbf{1.000} \\ \hline
\end{tabular}
\end{minipage}
\hfill
\begin{minipage}{0.485\textwidth}
\tablestyle{3.8pt}{1.2}
\caption{Ablation on pretrained weights for paper roll organization. ViVa with pretrained weights outperforms its non-pretrained counterpart.}
\label{tab:pretrain_paper}
\begin{tabular}{cccccc}
\hline
Method & Pretrained & MS & ES & MDR & ErrDR \\ \hline
ViVa w/o pretrain & \ding{55} & 0.034 & 0.034 & 0.489 & 0.667 \\
\rowcolor[HTML]{EFF6FB}
ViVa & \ding{51} & \textbf{0.039} & \textbf{0.055} & \textbf{0.762} & \textbf{0.778} \\ \hline
\end{tabular}
\end{minipage}
\end{table}

\paragraph{Effect of pretrained weights.}
We ablate the pretrained video weights on all three tasks, as reported in Tables~\ref{tab:pretrain_box}, \ref{tab:pretrain_shirt} and~\ref{tab:pretrain_paper}. Across the board, removing pretrained weights causes substantial degradation. On box packaging, MS drops from 0.034 to 0.009 and MDR falls from 0.733 to 0.523. The impact is even more severe on shirt folding, where ES plummets from $0.012$ to $-0.524$, effectively turning a reliable error detector into one that signals in the wrong direction at error events. On paper roll organization, MDR degrades from 0.762 to 0.489 and ErrDR drops from 0.778 to 0.667. These consistent results confirm that spatiotemporal priors from large-scale video pretraining are indispensable for building value models that generalize across tasks and reliably detect execution errors.

\subsection{Hyperparameter Analysis}
\label{sec:additional_exp}

\begin{figure*}[h]
    \centering
    \includegraphics[width=1.0\linewidth]{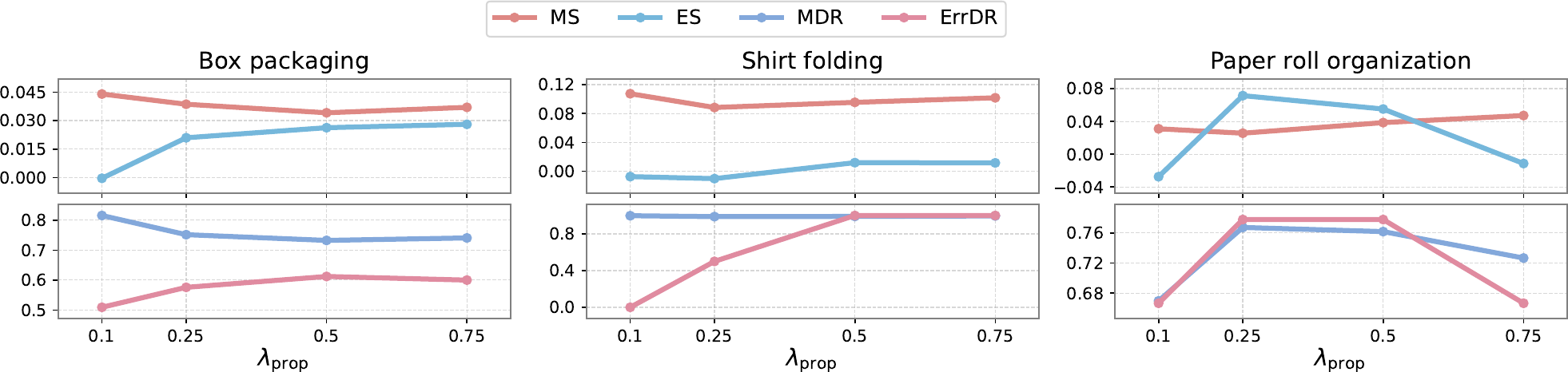}
    \caption{Loss weight analysis for balancing value and future proprioception predictions. Setting $\lambda_{\text{prop}}=0.5$ achieves the best trade-off and is adopted across all experiments.}
    \label{fig:ablate_loss}
\end{figure*}

\paragraph{Effect of loss weight.}
We vary \(\lambda_{\text{prop}}\) (the loss weight for future proprioception prediction) from 0.1 to 0.75, as reported in Figure~\ref{fig:ablate_loss}. Low values, such as \(\lambda_{\text{prop}}=0.1\), lead to negative error sensitivity (ES) on shirt folding and paper roll organization, and yield zero error detection (ErrDR) on shirt folding. This indicates that insufficient proprioception modeling weakens embodiment grounding for reliable error awareness. Increasing \(\lambda_{\text{prop}}\) to 0.5 yields positive ES across all three tasks and achieves high ErrDR (1.0 on shirt folding, 0.78 on paper roll organization, and 0.61 on box packaging). Further increasing to 0.75, however, degrades performance on paper roll organization (ES becomes negative again) and slightly reduces ErrDR on box packaging. Consequently, \(\lambda_{\text{prop}}=0.5\) offers the most robust and balanced performance across tasks and is adopted for all experiments.

\begin{figure*}[h]
    \centering
    \includegraphics[width=1.0\linewidth]{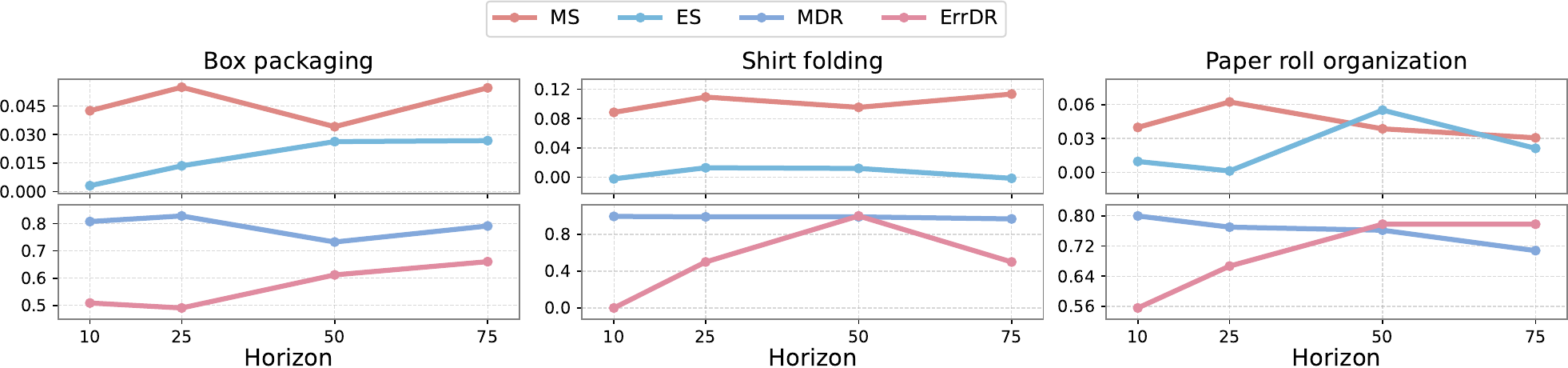}
    \caption{Hyperparameter analysis on the prediction horizon \(K\) for future proprioception prediction. Setting \(K=50\) achieves the best trade-off across tasks and is adopted for all experiments.}
    \label{fig:horizon_ablation}
\end{figure*}

\paragraph{Effect of prediction horizon.}
We vary the prediction horizon \(K\) across \(\{10, 25, 50, 75\}\), as shown in Figure~\ref{fig:horizon_ablation}. At \(K=10\), error detection collapses on shirt folding (ErrDR = 0.000, ES = \(-0.002\)), indicating that short horizons fail to capture sufficient temporal context. At \(K=25\), milestone sensitivity improves (MS reaches 0.055 on box and 0.062 on paper) but ES on paper remains near zero (0.001). At \(K=50\), ES becomes positive across all three tasks and ErrDR reaches 1.0 on shirt folding, achieving the best overall balance. At \(K=75\), MS peaks on shirt (0.113) but ES turns negative (\(-0.002\)) and ErrDR drops to 0.500, suggesting overly long horizons introduce noise. Thus \(K=50\) is adopted for all experiments.

\subsection{Efficiency Comparison}

Table~\ref{tab:comp_cost} compares the computational cost of three value model variants. The VLM-based baseline follows the lightweight design of $\pi_{0.6}^*$ but incurs the highest training cost of 6 GPU·days and inference latency of 0.32 seconds per frame, primarily due to its SigLIP~\citep{zhai2023sigmoid} visual encoder. The video-based variant predicting value alone achieves the fastest inference at 0.11 seconds and the lowest training cost of 3 GPU·days, yet omitting future proprioception compromises prediction accuracy. Our full ViVa model strikes a favorable balance, training in 4 GPU·days, 1.5× faster than the VLM baseline, while running at 0.18 seconds per frame. The additional proprioceptive prediction enriches the learning signal with minimal computational overhead.

\begin{table}[h]
\tablestyle{13pt}{1.4}
\caption{Computational cost comparison. Training time (GPU$\cdot$days) and inference time (s) are reported. ViVa achieves faster training and inference than the VLM-based baseline.}
\begin{tabular}{cccc}
\hline
Method & Backbone & Training cost (GPU$\cdot$days) & Inference cost (s / frame) \\ \hline
$\pi^{*}_{0.6}$ & VLM & 6 & 0.32 \\
ViVa w/o future-state & VG & 3 & 0.11 \\
ViVa & VG & 4 & 0.18 \\ \hline
\end{tabular}
\label{tab:comp_cost}
\end{table}

\section{Conclusion}
\label{sec:conclusion}
In this work, we introduced ViVa, a video-generative value model that grounds value estimation in predicted future dynamics. By jointly predicting future proprioception alongside value, ViVa learns temporally grounded representations that reliably track task progress and detect execution errors. ViVa achieves strong results across all three real-robot tasks, demonstrating the importance of spatiotemporal priors and embodiment-aware prediction for value learning.

\paragraph{Limitation.}
Due to resource constraints, we have not yet pretrained the value model at scale. We believe this would enable strong cross-task generalization, which we leave for future work.

\clearpage

\setcitestyle{numbers}
\bibliographystyle{plainnat}
\bibliography{main}
\end{document}